\documentclass{article}

\usepackage[nonatbib,preprint]{neurips_2022}

\usepackage{times}
\usepackage{epsfig}
\usepackage{graphicx}
\usepackage{amsmath}
\usepackage{amssymb}

\usepackage[utf8]{inputenc} 
\usepackage[T1]{fontenc}    
\usepackage{url}            
\usepackage{booktabs}       
\usepackage{amsfonts}       
\usepackage{nicefrac}       
\usepackage{microtype}      
\usepackage{xcolor}         

\usepackage{graphicx}
\usepackage[numbers]{natbib}
\usepackage{doi}

\begin{document}

\title{U(1) Symmetry-breaking Observed in \\ Generic CNN Bottleneck Layers}

\author{Louis-Fran\c{c}ois Bouchard ~~~ Mohsen Ben Lazreg ~~~ Matthew Toews\\ \\
{\'E}cole de Technologie Sup{\'e}rieure {\'E}TS\\
Montr{\'e}al, Canada\\
{\tt\small \{bouchard.lf, mohsenbenlazreg, matt.toews\}@gmail.com}



}

\maketitle

\begin{abstract}
We report on a novel model linking deep convolutional neural networks (CNN) to biological vision and fundamental particle physics. Information propagation in a CNN is modeled via an analogy to an optical system, where information is concentrated near a bottleneck where the 2D spatial resolution collapses about a focal point $1\times 1=1$. A 3D space $(x,y,t)$ is defined by $(x,y)$ coordinates in the image plane and CNN layer $t$, where a principal ray $(0,0,t)$ runs in the direction of information propagation through both the optical axis and the image center pixel located at $(x,y)=(0,0)$, about which the sharpest possible spatial focus is limited to a circle of confusion in the image plane. Our novel insight is to model the principal optical ray $(0,0,t)$ as geometrically equivalent to the medial vector in the positive orthant $I(x,y) \in R^{N+}$ of a $N$-channel activation space, e.g. along the greyscale (or luminance) vector $(t,t,t)$ in $RGB$ colour space. Information is thus concentrated into an energy potential $E(x,y,t)=\|I(x,y,t)\|^2$, which, particularly for bottleneck layers $t$ of generic CNNs, is highly concentrated and symmetric about the spatial origin $(0,0,t)$ and exhibits the well-known "Sombrero" potential of the boson particle. This symmetry is broken in classification, where bottleneck layers of generic pre-trained CNN models exhibit a consistent class-specific bias towards an angle $\theta \in U(1)$ defined simultaneously in the image plane and in activation feature space. Initial observations validate our hypothesis from generic pre-trained CNN activation maps and a bare-bones memory-based classification scheme, with no training or tuning. Training from scratch using combined one-hot $+ U(1)$ loss improves classification for all tasks tested including ImageNet.
\end{abstract}

\section{Introduction}

A modern computer vision system involves image acquisition, computational processing and memory indexing. Research in the last decade has focused intensely  on optimizing computational processing, i.e. deep neural network classification of labelled image data~\cite{krizhevsky2012imagenet,sun2019optimizationsurvey}, and more rarely considers aspects of acquisition. Nevertheless, both optical capture systems and deep neural networks propagate visual information across time, and operate by concentrating visual information in spatial bottlenecks. Might there be unexplored analogies between the two systems leading to new insights and improved architectures?
\\

\begin{figure}
\begin{center}
   \includegraphics[width=1\textwidth]{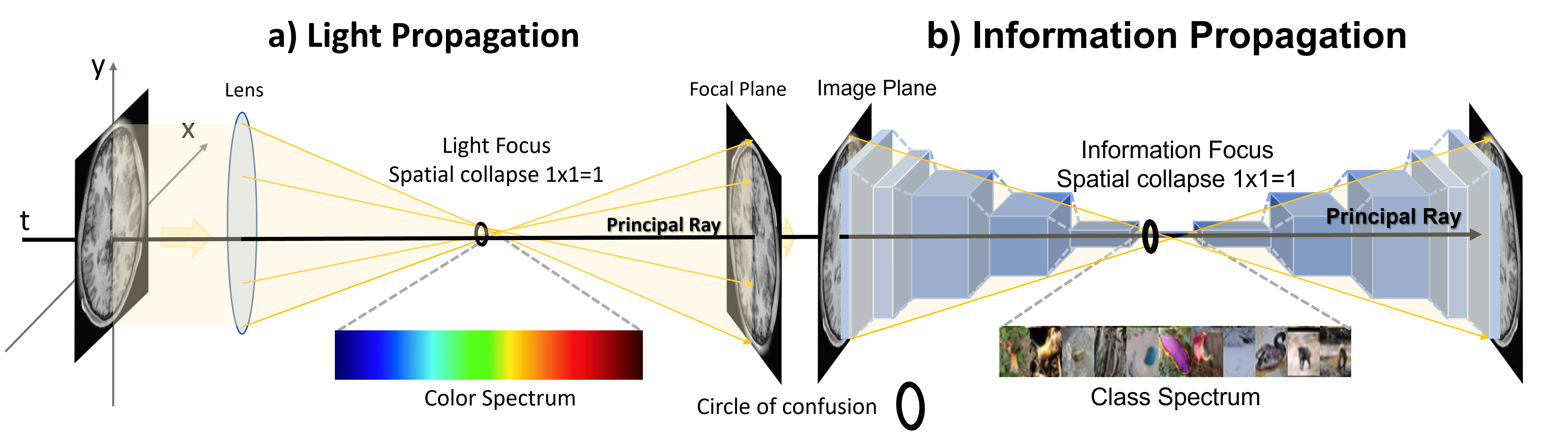}
\end{center}
   \caption{a) The propagation of light rays through an optical lens system. b) the propagation of information through a deep neural network (encoder-decoder). Shared aspects include a common principal ray, a spectrum and a circle of confusion.}
\label{fig:light_info_analogy}
\end{figure}

To answer this question, we propose to analyse information propagation through an optical system equipped with a lens and followed by a deep neural network encoder-decoder, as shown in Figure~\ref{fig:light_info_analogy}, noting several novel and remarkable analogies between both systems. First, multi-spectral information in the $(x,y)$ plane may be viewed as collapsing to a point $1\times 1=1$ resolution at both a) an optical focal point and b) a neural network bottleneck. These points are located on a ray $t$ perpendicular to the $(x,y)$ plane of both a) the propagating optical wavefront and b) network image layer. Information present in both $(x,y)$ spatial coordinates and activation channels $I(x,y)$ may thus be treated within a unified geometrical framework, centered about the principal ray. We pay special attention to the {\em circle of confusion}, which represents a minimum spatial radius within which information may be focused in an optical system, given lens imperfections including diffraction, lens aberrations, chromatic aberration of the input spectrum. Analogously, we hypothesize a circle of confusion limiting the maximum classification accuracy may be achieved in the spatial bottleneck of a deep neural network filtering process. Both observations and training experiments support a model whereby bottleneck activations for different classes are optimally distributed along a unit circle angle $\theta \in U(1)$ defined simultaneous in the image plane and in the activation feature space.

Our paper begins by presenting novel observations regarding the common structure of activation maps in bottleneck layers~\cite{zeiler2014visualizing, lenc2015understanding}, shared across diverse pre-trained architectures (ResNet~\cite{he2016deep}, DenseNet~\cite{huang2016densenet}, Inception~\cite{szegedy2016rethinking}, VGG~\cite{simonyan2015deep}) and datasets (textures, objects, fine-grained objects). In all cases, the activation energy $E(x,y)=\|I(x,y)\|^2$ is highly symmetric and concentrated near the image center $(x,y)$, highly reminiscent of the so-called "sombrero" potential associated with boson particles~\cite{goldstone1961field}. This symmetry is broken in classification trials via transfer learning from generic pretrained networks, via classic nearest-neighbor lookup~\cite{cover1967nearest} of pre-trained bottleneck activation information stored in memory. Lookup of individual pixel-wise activations rather than typical representations such pooled vectors or entire flattened layers both leads to superior classification accuracy and allows symmetry-breaking to be observed. Most notably, while matches between images of the same class are concentrated near the image center or origin $(x,y)=(0,0)$, they exhibit a consistent class-specific bias towards an angular direction $\theta \in U(1)$ in both the image plane and feature space. 

These observations lead us to investigate the hypothesis of a unitary class label $\theta \in U(1)$ parameter in network training from scratch. Training networks with a single random angular class label leads to improved classification across a wide variety of tasks including generic objects, textures, animals, and highly specific categories including bird species or aircrafts. We propose a combined $one-hot + U(1)$ loss function, introducing an antisymmetric training signal via an angular class label $(x, y) \in U(1)$ constrained to the unit circle $x^2 + y^2 = 1$, which leads to consistently higher accuracy across a variety of CNN architextures and classification tasks of varying specificity including ImageNet.

\section{Related Work}

Our work adopts an analogy between the propagation of light in an optical system and layer-wise CNN processing, where layers are evaluated in discrete time steps $t$. Deep neural networks may be used to simulate ordinary and partial differential equations~\cite{dissanayake1994neural,chen2018neural}, including solving forward and inverse problems and simulating wave propagation via graphics processing unit (GPU)-based CNN implementations~\cite{roska1995simulating,raissi2019physics}. This is unsurprising given that GPUs were designed initially for graphics algorithms such as ray tracing that approximate Maxwell's electromagnetic (EM) field equations for wavelengths much smaller than objects imaged (e.g. the $[380,750]$nm range of the visible RGB spectrum vs. human-sized objects). Neural networks may be computed via physical optical filters, where the difficulty is in achieving non-linear activations~\cite{sui2020review}. The key insight is that information may be modeled as propagating through space as a field evolving in discrete time steps, in a sequential, feed-forward manner with transitions defined by linear dot product operators $I_{t,\bar{x}}=I_{t-1,\bar{x}}\cdot W_{t,\bar{x}}$ in addition to non-linear activation, sub-sampling, pooling, etc. Our results derive from basic geometrical notions, the $\bar{x}=(x,y) \in R^2$ image space centered on $(x,y)=(0,0)$, a 3D spacetime $(t,\bar{x}) \in R^+\times R^2$, and a vector-valued activation image $I(t,\bar{x}) \in R^N$. Our novel approach considers a single unitary $U(1)$ variable, i.e. $(x,y)$ coordinates constrained to the unit circle $x^2+y^2=1$ or equivalently an angle $\theta = atan2(y,x)$. $U(1)$ may be also understood by considering that any function $f(\bar{x})$ may be expressed as the sum $f(\bar{x})=f_s(\bar{x})+f_a(\bar{x})$ of symmetric $f_s(\bar{x})$  and anti-symmetric $f_a(\bar{x})$ components, i.e. the $cos(\bar{x})$ and $sin(\bar{x})$ components of Euler's equation $e^{i\theta}=\cos(\theta)+i\sin(\theta)$. Unitary variables are well known in mathematical analysis and increasingly used in machine learning formulations~\cite{kiani2022projunn,tang2021image}. In the standard model of 3D particle physics, fundamental particles are defined as fields or wave functions $\psi(\bar{x})$ referred to as bosons $\psi_s(\bar{x})$ if the wave function is symmetric (eg. notably the photon and the Higgs boson) and as fermions $\psi_a(\bar{x})$ if it is anti-symmetric (e.g. quarks and leptons, notably the electron).

Symmetry-breaking is a process by which an energy potential transitions from a symmetric to an asymmetric state. In deep neural networks, early investigations included replica symmetry breaking~\cite{monasson1994domains} and non-monotonic neural networks~\cite{boffetta1993symmetry}, more recently Euclidean neural networks~\cite{smidt2021finding} and Noether's notions of continuous symmetry from Lagrangian energy formulations ~\cite{tanaka2021noether}.
In biological networks it has been notably investigated as an orientation selection mechanism within the cortex~\cite{ben1995theory}, ring attractor dynamics projecting individual photons into the fruit fly brain and measuring the angular response~\cite{kim2017ring}. In particle physics, the Higgs mechanism lending mass to other standard particles is based on symmetry breaking, first proposed in 1962~\cite{higgs1964broken} and experimentally confirmed in 2012~\cite{aad2012observation} (Nobel prize 2013), the same year AlexNet ushered in the wave of GPU-trained deep neural networks~\cite{krizhevsky2012imagenet}. Most recently, 2D quasi-particles called anyons were theorized in 1982~\cite{wilczek1982quantum} and observed experimentally in 2020 as electrons constrained to 2D sheets of gallium arsenide~\cite{bartolomei2020fractional}. Anyon wavefunctions are not constrained to be exclusively symmetric or anti-symmetric, as are standard 3D particles (e.g. photons or electrons), and offer a potential mechanism for quantum computing and memory formation via a braiding-like particle exchange process in the 2D plane.

Our initial observations of $U(1)$ symmetry-breaking are based on nearest neighbor indexing and classification~\cite{cover1967nearest}, specifically using spatially-localized activation vectors in pre-trained networks. This follows the transfer learning approach, where networks pre-trained on large generic datasets such as ImageNet~\cite{jdeng2009imagenet} are used as general feature extractors for new tasks~\cite{kornblith2019better,azizpour2015factors,cimpoi2016deep}. Deep bottleneck activations tend to outperform specialized shallower networks and meta-learning methods~\cite{chen2019closer},  particularly in the case of few training data and a large domain shift between training and testing data~\cite{guo2020broader}. Various approaches seek to adapt ImageNet models to fine-grained tasks by encoding activations at bottleneck layers, e.g. via descriptor information (e.g. extracted off-the-shelf features~\cite{sharif2014cnn}, VLAD~\cite{arandjelovic2016netvlad}), global average or max pooling~\cite{razavian2016visual}, generalized mean (GeM)~\cite{radenovic2018fine}, regional max pooling (R-MAC)~\cite{tolias2016particular} in intermediate layers, modulated by attention operators~\cite{noh2017large}. Additional training may consider joint loss between classification and instance retrieval terms~\cite{berman2019multigrain}. The mechanism of spatially localized activations (as opposed to global descriptors) is closely linked to the attention mechanisms~\cite{huang2019ccnet}, including non-local networks~\cite{wang2018non}, squeeze-and-excitation networks~\cite{hu2018squeeze},
transformer architectures
\cite{vaswani2017attention,carion2020end,han2020survey} including hierarchically shifted windows~\cite{liu2021swin}, thin bottleneck layers~\cite{sandler2018mobilenetv2}, self-attention mechanisms considering locations and channels~\cite{woo2018cbam}, intra-kernel correlations~\cite{haase2020rethinking}, multi-layer perceptrons incorporating Euler's angle~\cite{tang2021image}, correspondence-based transformers~\cite{jiang2021cotr} and detectors~\cite{sun2021loftr}, and geometrical embedding of spatial information via graphs~\cite{kipf2016semi, henaff2015deep}. Whereas these works typically seek end-to-end learning solutions fitting within GPU memory constraints~\cite{gordo2016deep}, we seek to demonstrate the $U(1)$ theory via basic memory lookup. 

Our final results training from scratch using $U(1)$ labels is similar in spirit to work seeking to regularize the label and/or the activation feature space, including using real-valued rather than one-hot training labels~\cite{rodriguez2018beyond}, learning-based classifiers~\cite{yanwang2019simpleshot,wen2016discriminative}, prototypical networks for few-shot learning~\cite{nguyen2020sen,snell2017prototypical}, deep k-nearest neighbors~\cite{papernot2018deep}, geometrical regularization based on hyperspheres~\cite{mettes2019hyperspherical, shen2021spherical}, enforcing constant radial distance from the feature space origin~\cite{zheng2018ring} or angular loss between prototypes~\cite{wang2017deep}. We seek to present our theory in the broadest, most general context. We consider a basic classification and deliberately eschew architectural modifications that might limit the generality of our analysis. We demonstrate our general theory with results using a wide variety of pre-trained architectures including DenseNet~\cite{huang2016densenet}, Inception~\cite{szegedy2016rethinking}, ResNet~\cite{he2016deep}, VGG~\cite{simonyan2015deep} directly imported from TensorFlow~\cite{tensorflow2015-whitepaper}. We consider a variety of testing datasets not used in ImageNet~\cite{jdeng2009imagenet} training, with including general categories (e.g. Caltech 101~\cite{feifei2006caltech}), and specific instances (e.g. human brain MRIs of family members~\cite{VANESSEN201362}, faces~\cite{zhang2017utkface}).


\section{Information Propagation and $U(1)$ Symmetry-breaking}
\label{3. Information Propagation}
We propose modeling information propagation through a generic visual information processing system, including both the optical capture apparatus and discrete layer-wise processing in a deep convolutional neural network, as shown in Figure~\ref{fig:light_info_analogy}. We develop an analogy between both, in order to develop the notion of an information spectrum, unit circle of confusion and generic classification as a $U(1)$ symmetry-breaking process.

The flow of information through spacetime in a computer vision system may be viewed as the propagation of information quanta through spacetime. In an optical system as in Figure~\ref{fig:light_info_analogy} a), information is carried by photons radiating from a 3D scene, where photons travel along rays in 3D space at a constant speed in a vacuum. Light rays concentrated at a focus via a lens or a pinhole may be accumulated into a spatially coherent 2D image at an image plane, e.g. a RGB photosensor array. In a feedforward neural network system as in Figure~\ref{fig:light_info_analogy} b), information is carried by activation vectors and propagating layer-wise operations including linear dot product, non-linear activation functions such as ReLu~\cite{vinod2010reluboltzmann}, pooling and subsampling. Information concentrated at a focus may be used for prediction or propagated further to achieve a spatially coherent 2D image segmentation~\cite{long2018segm}.

Several aspects common to both systems may be noted. Information is distributed across a 2D spatial plane $(x,y)$, i.e. a propagating light wavefront or a deep network layer, and across a spectrum $I(x,y)$, i.e. the tri-chromatic RGB spectrum of visible light or the activation channels of a network layer. We define the principal ray as perpendicular to the $(x,y)$ image plane, and {\em central to both the 2D image space and information channel space}. In both systems, information is routed towards a focal point along the principal ray, where the finest achievable focus is limited to a circle of confusion, shown as a black ring in Figure~\ref{fig:light_info_analogy}. The circle of confusion is a well-known optical phenomenon arising from diffraction or lens imperfections, and compounded by dispersion, channel-specific refraction and chromatic aberration. We hypothesize a similar phenomenon in deep neural networks, due to the point spread function of filtering operations, and this motivates our investigation into bottleneck disentanglement.

In an optical system, the electromagnetic (EM) spectrum encodes information as photon particles, which may be modeled as propagating through 3D space via Dirac delta operators $\delta(d\bar{x})$ (i.e. translation filters) defined in 3D space by the geometry of the optical apparatus. Photons are so-called boson particles defined by a symmetric wave function $\psi(x)=\psi_s(x)=\psi_s(-x)$, and may accumulate at a spatial focus according to Bose-Einstein statistics~\cite{bose1924plancks}~\footnote{As opposed to so-called fermion particles such as electrons defined by an anti-symmetric wave function $\psi(x)=\psi_a(x)=-\psi_a(-x)$ which may only accumulate in (spin-up, spin-down) pairs due to the Pauli exclusion principle.}. In a deep neural network, information is entangled in a limited number of activation channels $I(x,y)\in R^{N+}$, and propagated via multi-channel filtering operations. An important difference between the two systems is that in the case of deep CNNs~\cite{lecun1989backprop}, the channel spectrum or gauge~\cite{cohen2019gauge} is transformed at each layer in order to encode and compress increasing amounts of visual information at reduced spatial resolutions up to the network bottleneck, trading equivalent representations for class-specific information the deeper we get into the network~\cite{zeiler2014visualizing, lenc2015understanding}, whereas the EM gauge is preserved during light propagation.

The propagation of information in both systems may be modeled as a vector field propagating through $(x,y,t)$ spacetime via local dot product operators as follows. Let $I_{t,\bar{x}}$ be a vector of multi-channel information at layer $t$ and spatial location $\bar{x}=(x,y)$, for example a 3-channel RGB image or a N-channel activation layer. Information may be propagated from time $t$ to $t+1$ by the dot product operation $I_{t+1,\bar{x}} = \bar{I}_{t,\bar{x}} \cdot W_{t,\bar{x}}$, where $\bar{I}_{t,\bar{x}}$ is a neighborhood tensor defined in the spatial window surrounding $(t,\bar{x})$, and $W_{t,\bar{x}}$ are the weights of a linear filter of the same size and shape as $\bar{I}_{t,\bar{x}}$. Note that in a convolutional neural network (CNN)~\cite{lecun1989backprop}, filters are translation invariant and constant across a layer $W_{t,\bar{x}}=W_t$~\cite{zeiler2014visualizing, lenc2015understanding}, however they may generally vary with spatial location as in multi-layer perceptrons~\cite{Rosenblatt58theperceptron} or transformer networks~\cite{vaswani2017attention} (hence transformers 'doing away with convolution').



\begin{figure}[ht]
\begin{center}
   \includegraphics[width=.8\linewidth]{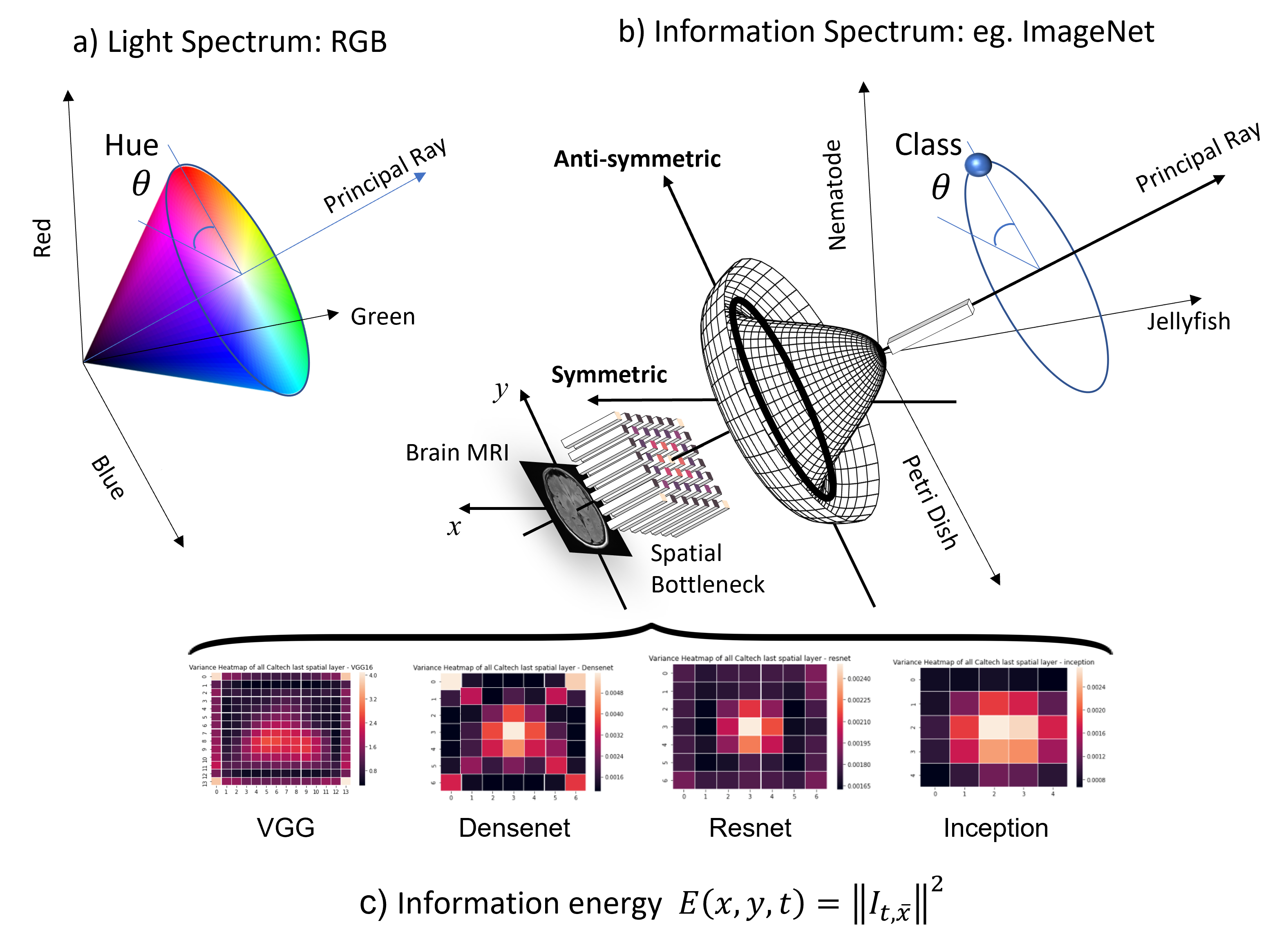}
\end{center}
   \caption{Our polar coordinate model for multi-channel data. a) The standard 3-channel red-green-blue color space with hue angle $\theta$. b) A 3-channel subspace of the 1000-channel ImageNet~\cite{jdeng2009imagenet} output, following a spatial bottleneck layer, showing a class angle $\theta$ analogous to color hue and defined by the input image content. (c) Information energy maps $E_{t,\bar{x}}=\|I_{t,\bar{x}}\|^2$ computed from spatial bottleneck layers of various ImageNet pre-trained networks and across Caltech101~\cite{feifei2006caltech} classes not used in training. Note the highly symmetric structure and similarity to the well-known bosonic "sombrero" energy potential~\cite{goldstone1961field}. b) shows how a specific image (e.g. here a human brain MRI class not found in ImageNet) exhibits asymmetric bias towards an angle $\theta$ in the direction of visually similar ImageNet classes axes (here Nematode, Petri Dish and Jellyfish).}
\label{fig:hsv_csv}
\end{figure}

Our model is illustrated in Figure~\ref{fig:hsv_csv}, where multi-channel activation information is non-negative (e.g. following rectification (ReLu)~\cite{vinod2010reluboltzmann}), located in the positive orthant of activation space $I_{t,\bar{x}} \in R^{N+}$, and may be expressed in polar coordinates $(r,\theta,t)$ centered about the spatial origin $(x,y)=(0,0)$ and along the principal ray $(0,0,t)$. For example, RGB color pixels may be represented by hue, saturation, intensity (HSV) coordinates, where color perception is closely linked to a hue angle $\theta \in U(1)$. We propose an analogy whereby multi-channel CNN activations are also represented via a {\em class angle} $\theta \in U(1)$ according to dominant intensity channels. We hypothesize that this is a primary mechanism by which class information is encoded in image space, and that it may be identified and used as a training signal in the $(x,y)$ image plane of arbitrary CNN layers, most notably bottleneck layers with minimal spatial extent. We observe the scalar energy $E_{t,\bar{x}}=\|I_{t,\bar{x}}\|^2 \in R^+$ of activation layers $I_{t,\bar{x}}$ from a variety of generic architectures pretrained on the ImageNet dataset~\cite{jdeng2009imagenet} in response to input images not used in training (e.g. Caltech 101~\cite{feifei2006caltech}). As shown in Figure~\ref{fig:hsv_csv}, the average activation energy of pre-trained bottleneck layers is generally distributed symmetrically about the origin in a "sombrero" shape reminiscent of a bosonic wave function. This symmetry is broken in classification as shown in Figure~\ref{fig:hsv_csv} b), class-specific activation information exhibits a consistent asymmetric bias towards an angle $\theta \in U(1)$ shared across individuals of a class.


\section{Experiments}

Our experiments validate our $U(1)$ theory in two sections. We first demonstrate symmetry-breaking in basic nearest neighbor classification experiments, for a variety generic ImageNet pre-trained CNNs. We then perform training with $U(1)$ class labels, using a novel one-hot + $U(1)$ loss function modeling symmetric + antisymmetric components.

\subsection{Observing Symmetry-breaking in CNN Bottleneck Layers}

\begin{figure}
\begin{center}
   \includegraphics[width=1\linewidth]{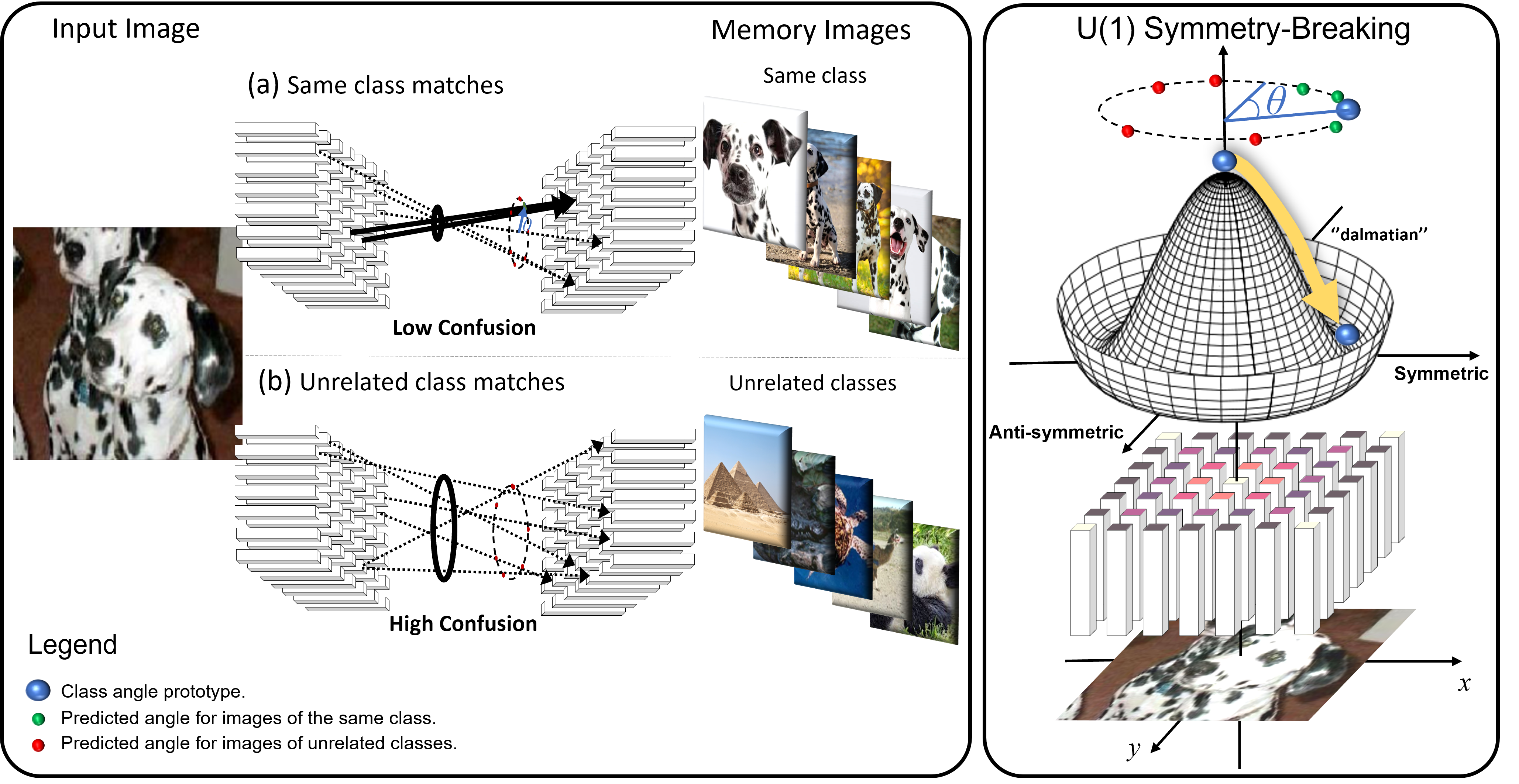}
\end{center}
   \caption{(Left) Our proposed retrieval architecture where classification is achieved from nearest neighbor matches between activation vectors derived from an input image and labeled images stored in memory. Matches between images of the same class (a) generally exhibit lower spatial confusion than (b) matches to unrelated class images. (Right) Our hypothesis that image-to-memory matching leads to symmetry-breaking of the layer energy function over the $U(1)$ plane and class recognition.}
\label{fig:memory_transformer}
\end{figure}

We hypothesize that class information in trained networks is generally concentrated into an angular argument $\theta \in U(1)$ both in the $(x,y)$ image plane and in activation feature space. We begin with observations supporting this hypothesis, where classification via basic nearest-neighbor indexing of pre-trained network activations is achieved via a symmetry-breaking process in the image plane. Figure~\ref{fig:memory_transformer} illustrates our proposed experimental retrieval architecture, where individual vectors $I_{t,\bar{x}}$ may translate across image space in order to match with similar vectors in memory extracted from similar classes. As noted, the general mechanism is closely linked to spatial attention approaches, and our implementation is designed to maximize baseline classification performance. Note also that minimizing the distance between vectors is equivalent to maximizing their dot product.

Classification is achieved by maximising the likelihood function $p(I|C,\{I'\} )$ of class $C$ associated with input image $I$ from set of image examples $\{I'\}$ stored in memory:
\begin{align}
    C^* = \underset{C}{\mathrm{argmax}}~p(I|C,\{I'\}),
\end{align}
where $C^*$ is a maximum likelihood (ML) estimate of the image class. The likelihood function is defined as follows. Let $i$ and $j$ be indices applying respectively to query and memory images. Let $x_i,x_j \in R^2$ be discrete spatial point locations, $x_i \in \Omega_{I}$ in the query image and $x_j \in \Omega_{I'}$ in memory images. Indices $j\in NN_i$ belong to a set of K nearest neighbors of associated with $i$ defined as $NN_i: \{j: \|I_{xi}-I'_{xj}\| \le \|I_{xi}-I'_{xk}\| \}$ identified via activation vector memory indexing, where $I'_{xk}$ is the $k^{th}$ nearest neighbor of $I_{xi}$ in memory. The likelihood may be expressed as a kernel density
\begin{align}
p(I|C,\{I'\} )  \propto
\frac{ \sum_{i}\sum_{j \in N_i} f(I_{x_i}-I'_{x_j}) [C=C'_j]}{ \sum_{x_j} [C=C'_j]},
 \label{eq:likelihood}
\end{align}
where $[C=C'_j]$ is the Iverson bracket evaluating to 1 upon equality and 0 otherwise, and the denominator normalizes for class frequency across the entire memory set $\sum_{j} [C=C'_j]$. The kernel function in Equation~\eqref{eq:likelihood} is based on activation vector (dis)similarity $f(I_{x_i},I'_{x_j})$ and is defined as:
\begin{align}
f(I_{x_i},I'_{x_j}) = exp- \left\{ \frac{ \| I_{x_i} - I'_{x_j} \|^2  }{\alpha_i^2+\epsilon } \right\},
\label{eq:activation_kernel}
\end{align}
where in Equation~\eqref{eq:activation_kernel}, $\alpha_i = \underset{x_j \in \Omega_{I'} }{\operatorname{min}} \| I_{x_i}-I'_{x_j}\|$ is an adaptive kernel bandwidth parameter defined as the distance to nearest activation vector $I_{x_i}$ in memory $I'_{x_j} \in \{I'\}$, and $\epsilon$ is a small positive constant ensuring a non-zero denominator. Note that the Euclidean and cosine distances are equivalent $\operatorname{argmin} \| I_{x_i}-I'_{x_j}\| = \operatorname{argmin} \{ 1-I_{x_i} \cdot I'_{x_j} \}$ for magnitude-normalized descriptors $\|I_{x_i}\|= \|I'_{x_j}\|=1$, and may be used interchangeably according to the computational architecture at hand.

\begin{figure}[ht]
\begin{center}
   \includegraphics[width=1\linewidth]{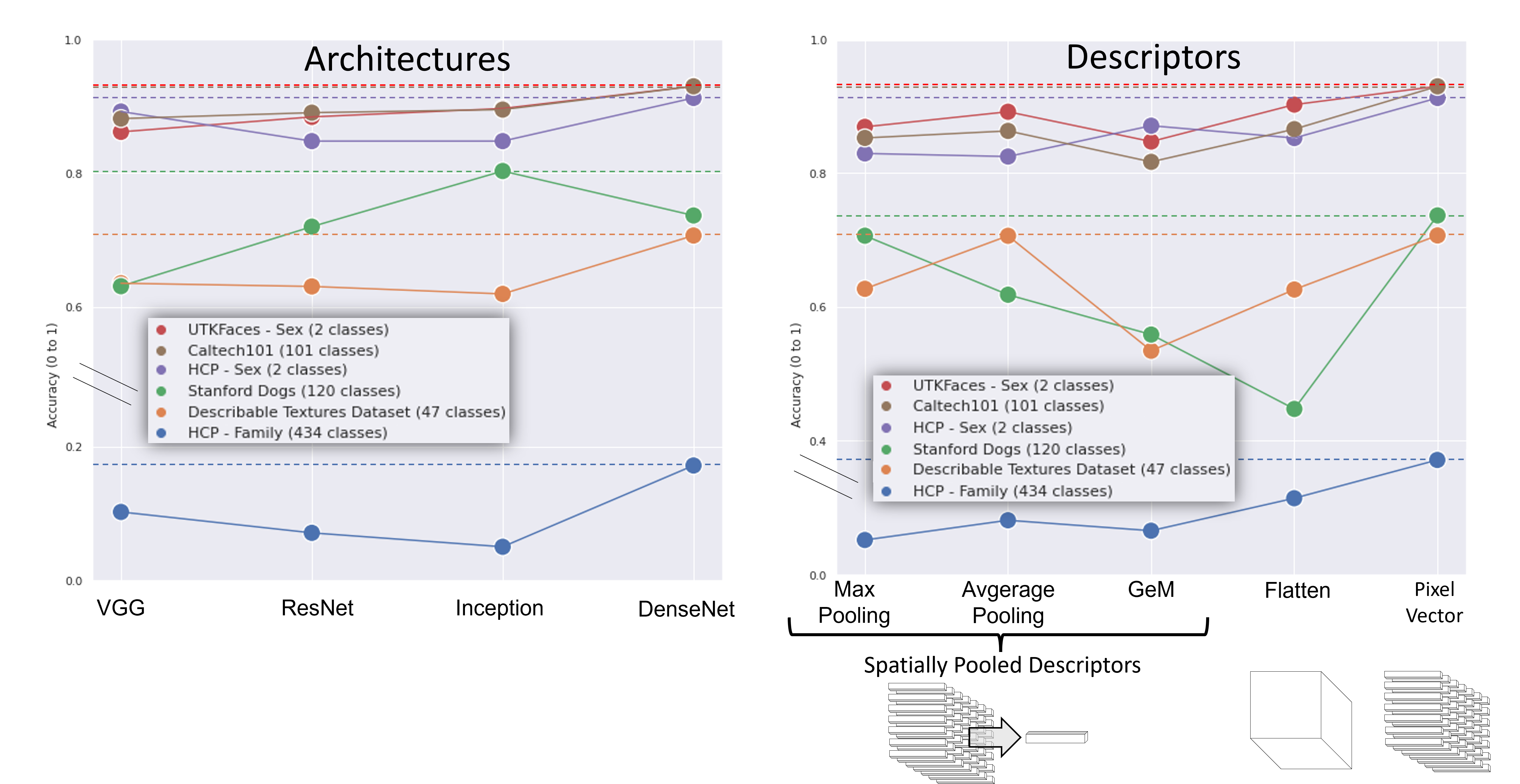}
\end{center}
   \caption{Baseline classification performance of architectures and descriptors on six memory-based classification tasks with $(K=10)$ nearest neighbors. Dashed lines indicate the superior performance of the Densenet~\cite{huang2016densenet} architecture (max in 5 of 6 tasks) and pixel vector descriptors (max for all).}
\label{fig:descriptor_comparison}
\end{figure}

Experiments are based on a variety of generic CNN architectures trained on the ImageNet dataset~\cite{jdeng2009imagenet}, and tested in basic memory-based retrieval experiments using various datasets specifically not used in training, including general objects (Caltech 101~\cite{feifei2006caltech}), textures (DTD)~\cite{cimpoi14describing} and specific objects~\cite{VANESSEN201362, zhang2017utkface}. Figure~\ref{fig:descriptor_comparison} establishes baseline classification results across CNN architectures and descriptors, showing that localized pixel vector matching lead to the highest accuracy amongst alternatives, particularly for Densenet~\cite{huang2016densenet}. Note that high accuracy for pixel vector matching is due to the trade-off between memory and computation (e.g. 7x7=49 pixels vs. 1 global descriptor). As our goal here is to observe symmetry-breaking via the most accurate spatially localized information available, however, efficiency is not an immediate concern and is beyond the focus of our work here. We achieve efficient retrieval using the Approximate Nearest Neighbor library (Annoy) method~\cite{annoy2016} indexing method, a rapid tree-based algorithm with $O(log~N)$ query complexity for $N$ elements in memory. Further efficiency could be achieved via compression (e.g PCA~\cite{pearson1901pca}) or specialized architectures~\cite{sun2021loftr,jiang2021cotr}.

\begin{figure}[ht]
\begin{center}
   \includegraphics[width=1.0\linewidth]{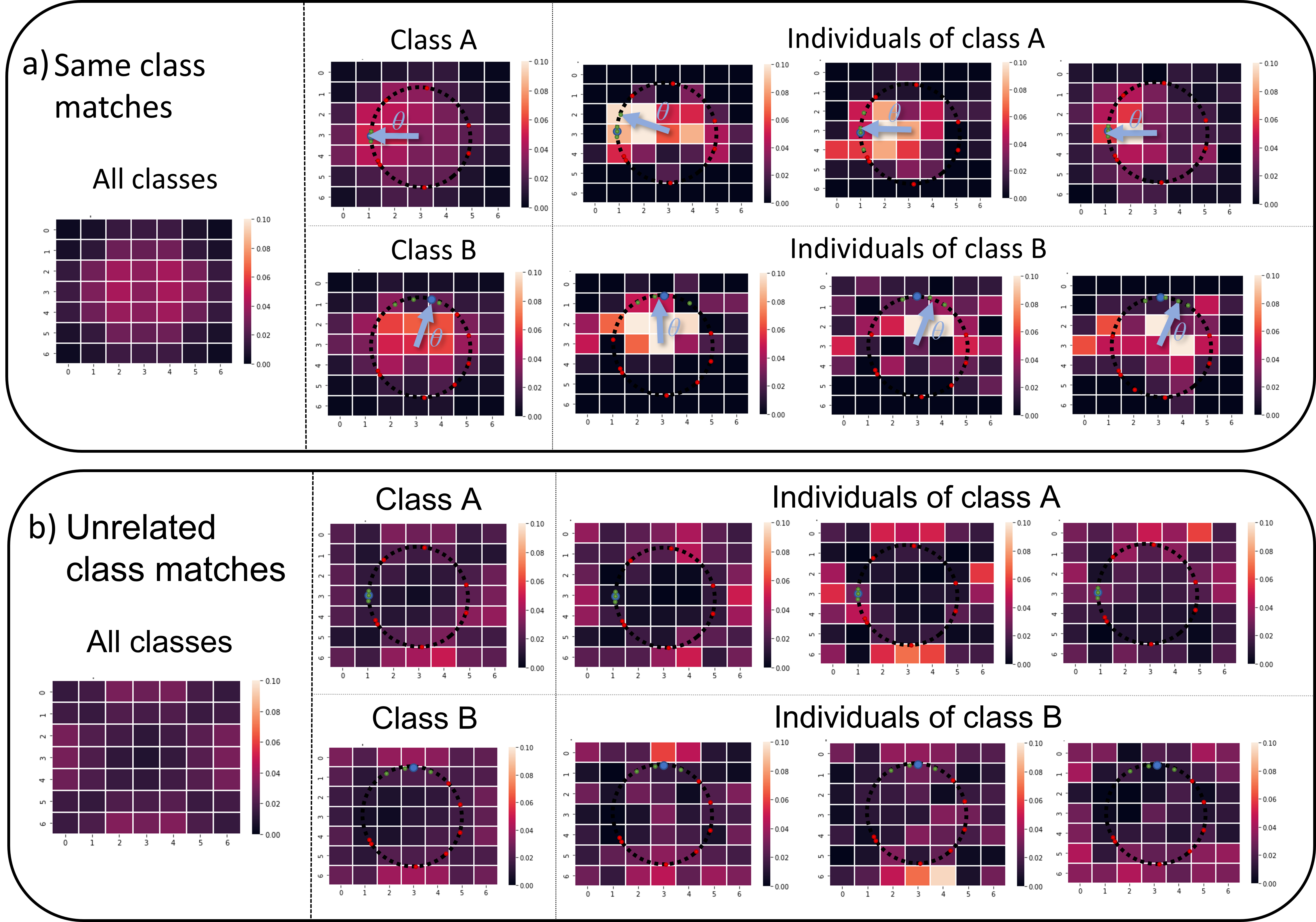}
\end{center}
   \caption{$U(1)$ symmetry-breaking observed in distributions of activation vector matches for Caltech 101~\cite{feifei2006caltech} classes. a) Matches between images of the same class are concentrated within a circle of confusion around the center $(0,0)$, and exhibit consistent bias towards a similar angle $\theta$ for specific classes and individuals of the same class (eg. A and B, blue arrows). b) Matches between images of unrelated class are scattered about the periphery.}
\label{fig:pixelmatching_pxi}
\end{figure}

Examples of consistent $U(1)$ symmetry-breaking are shown in Figure~\ref{fig:pixelmatching_pxi}. Note how distributions of nearest neighbor pixel vector matches between images of same class are consistently biased towards a similar angle $\theta$, validating class-specific asymmetry as predicted by our model in Figure~\ref{fig:hsv_csv}. Visualizations in Figure~\ref{fig:pixelmatching_pxi} make use of the $7 \times 7=49$ pixel DenseNet201~\cite{huang2016densenet} architecture with $I_{t,\bar{x}} \in R^{1920}$ channel vectors, which led to the highest accuracy in memory-based classification trials. 

\subsection{Training with One-hot + $U(1)$ Loss}


Given the pattern of regular angular symmetry-breaking observed in the case of pre-trained network indexing, we hypothesize that orientation may be learned from arbitrary $U(1)$ class labels in order to improve classification. Training combines a standard one-hot loss function with an additional angular $U(1)$ loss, and applies to generic networks and training from scratch with no bells or whistles. $U(1)$ loss uses a single angular training label, selected randomly for each class on a unit circle $\theta \in U(1)$ in the $(x,y)$ plane. $U(1)$ labels are parameterized as $(x,y)$ coordinates constrained to $x^2+y^2=1$, and estimated via fully-connected layers immediately following the bottleneck. The L2 distance between $U(1)$ labels and predicted $(x,y)$ parameters is used as a loss function, and mixed in equal weighting with one-hot difference to generate a combined cross-entropy loss for the backward step. We train over a fixed 100 epochs, with original train, validation, and test sets and 5-fold cross-validation on a single Titan RTX GPU. We use the Adam optimizer~\cite{adam} and CosineAnnealingLR scheduler~\cite{loshchilov2016sgdr} with default parameters from PyTorch~\cite{NEURIPS2019_9015} in all experiments and basic data augmentation in the form of horizontal flips and random crops. Training and classification are evaluated in diverse few-shot learning tasks of varying degrees of granularity, including the Describable Textures Dataset (DTD)~\cite{cimpoi14describing}, Caltech-UCSD Birds~\cite{399birds}, Stanford Dogs~\cite{KhoslaYaoJayadevaprakashFeiFei_FGVC2011}, Flowers~\cite{Nilsback08}, Pets~\cite{parkhi12a}, Indoor67~\cite{Quattoni2009recogscene}, FGVC-Aircraft~\cite{maji13fine-grained}, Cars~\cite{KrauseStarkDengFei-Fei_3DRR2013}, and Imagenet~\cite{jdeng2009imagenet}. 

\begin{figure}[ht]
\begin{center}
   \includegraphics[width=.7\linewidth]{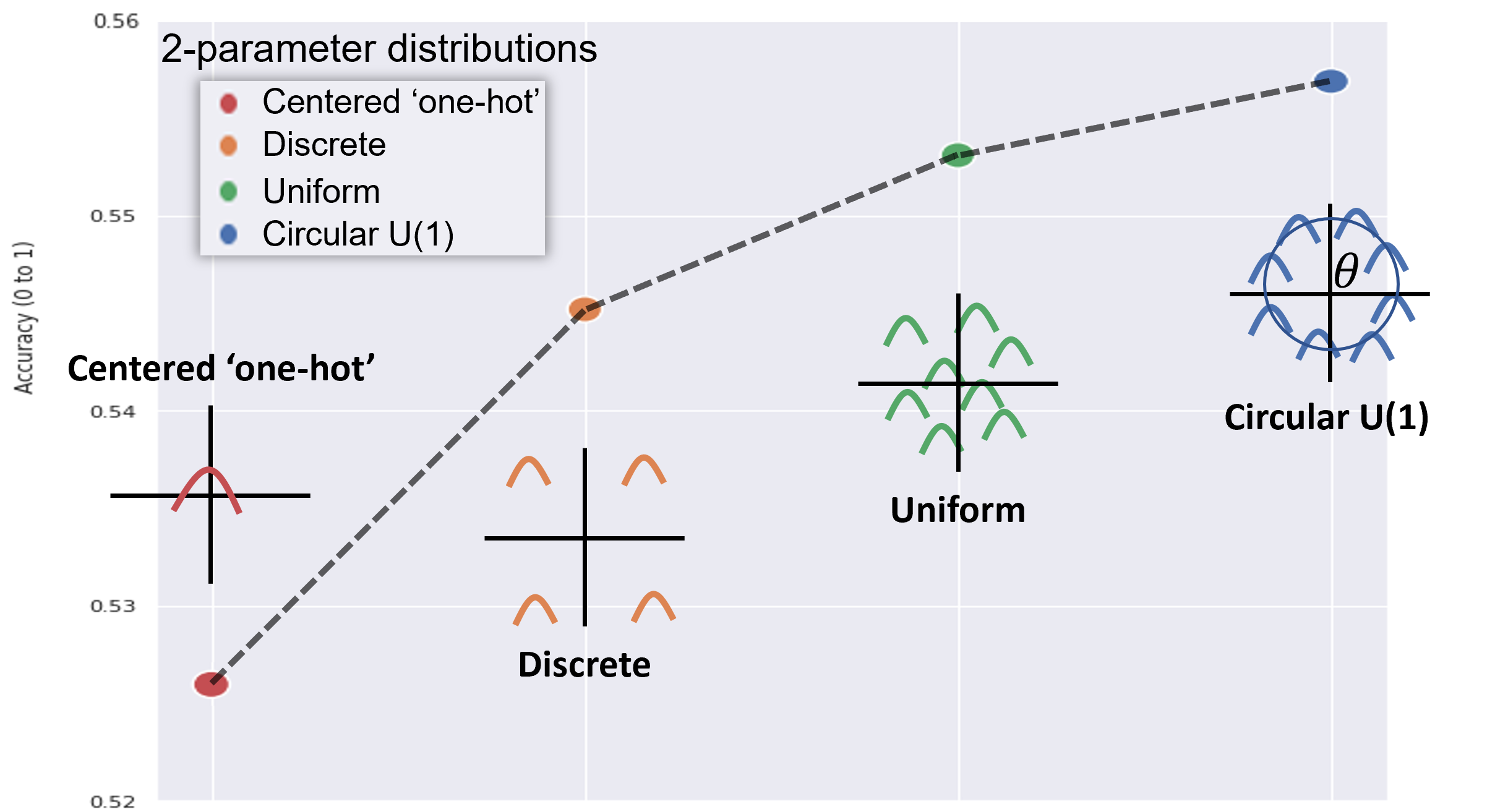}
\end{center}
   \caption{Comparing classification accuracy for training from scratch with various 2-parameter $(x,y)$ label configurations: one-hot, discrete, uniform and unit circle $U(1)$. Our proposed $U(1)$ leads to the highest accuracy (here using the Caltech-UCSD Birds dataset~\cite{399birds}).}
\label{fig:ucsd-tests}
\end{figure}

Training with one-hot + $U(1)$ loss improved classification for all tested networks compared to one-hot alone. Table~\ref{table:angular_results} reports results for the network architectures leading to the highest overall accuracy (EfficientNet-B0)~\cite{mingxing2019efficientnet} and the most improved (Resnet)~\cite{he2016deep}. As an additional experiment, we tested $(x,y)$ labels generated from various 2-parameter distributions other than the unit circle $U(1)$, including centered (one-hot alone), discrete binary combinations $[\pm 1,\pm 1]$ and uniformly distribution over space. Figure~\ref{fig:ucsd-tests} shows preliminary results for a single dataset, where $U(1)$ labels lead to the highest accuracy amongst alternatives.

\begin{table}[ht]
\resizebox{1\textwidth}{!}{%
\begin{tabular}{p{3.2cm}|p{1.2cm} p{1.2cm} p{1.2cm} p{1.5cm} p{1.1cm} p{1.7cm} p{1.6cm} p{1cm} p{1.2cm}}
 
    Model & DTD~\cite{cimpoi14describing} & UCSD Birds~\cite{399birds} & Stanford Dogs~\cite{KhoslaYaoJayadevaprakashFeiFei_FGVC2011} & Flowers~\cite{Nilsback08} & Pets~\cite{parkhi12a} & Indoor67~\cite{Quattoni2009recogscene} & FGVC-Aircraft~\cite{maji13fine-grained} & Cars~\cite{KrauseStarkDengFei-Fei_3DRR2013} & Imagenet~\cite{jdeng2009imagenet} \\ 
    \hline
        \textbf{ResNet-18}~\cite{he2016deep}\\ 
        One-hot & 0.4153 & 0.3875 & 0.4015 & 0.4863 & 0.4090 & 0.4866 & 0.8703 & 0.3575 & 0.6350 \\ 
        One-hot + $U(1)$ & \textbf{0.5230} & \textbf{0.5313} & \textbf{0.5754} & \textbf{0.6487} & \textbf{0.5919} & \textbf{0.6224} & \textbf{0.8920} & \textbf{0.8011} & \textbf{0.6621} \\
        \hline
        \textbf{EfficientNet-B0}~\cite{mingxing2019efficientnet}\\ 
        One-hot & 0.5310 & 0.4907 & 0.5655 & 0.7252 & 0.6689 & 0.5458 & 0.8685 & 0.7882 & 0.7144 \\
        One-hot + $U(1)$ & \textbf{0.5368} & \textbf{0.5569} & \textbf{0.5764} & \textbf{0.7438} & \textbf{0.6797} & \textbf{0.5477} & \textbf{0.8768} & \textbf{0.8079} & \textbf{0.7171} \\
    
\end{tabular}
  }
\caption{Classification results using training based on $U(1)$ labels.}
\label{table:angular_results}
\end{table}

\section{Discussion}

We present a novel model of information propagation in deep CNN layers with an analogy to light propagation in an optical system, including a focal point and an inherent circle of confusion in both cases. We assume the equivalence of the principal optical axis $(0,0,t)$ and the medial vector of positive activation feature space. These are both central to the $(x,y)$ image plane allow class to be characterized by a hue-like angle $\theta \in U(1)$, defined both in the image plane and activation feature space. Our model provides an 'end-to-end' view including also the optical capture system and memory storage. Our most significant finding pertains to information regarding $(x,y)$ image space, which in its most concentrated form in low-resolution network layers just prior to spatial collapse $1\times 1=1$, is well-represented by a single unitary variable $\theta \in U(1)$. Furthermore, the process by which classification occurs involves $U(1)$ symmetry breaking of the deep CNN bottleneck energy $E(x,y)=\|I_{t,\bar{x}}\|^2$, consistent with the "sombrero" bosonic potential. Our observations of a regular angular asymmetry consistent with image class are the first in the literature, to our knowledge, and indicate that symmetry-breaking occurs naturally in generic pre-trained CNNs, linking deep learning more closely to biological networks and recent results in particle physics. Our angular observations are consistent with biological models: specifically $\theta$ deviations in Figure~\ref{fig:pixelmatching_pxi} induced from image inputs, and $\theta$ deviations in orientation/heading selection in the fruit fly visual system induced by individual photon inputs, see~\cite{kim2017ring} Figure 3. They are also reminiscent recently discovered anyon quasiparticles~\cite{bartolomei2020fractional}, and may provide mechanism for memory formation whereby an input sequence is woven into a "braid" of memory via the particle exchange process in the 2D plane.



Our work brings together a broad range of fundamental concepts and state-of-the-art research in optics, particle physics, biological and artificial neural networks, which we attempt to present in the broadest and most understandable description, avoiding as much as possible domain-specific jargon. Our experiments make use of the most widely-known, generic neural network architectures, data sets and NN classification~\cite{cover1967nearest} in order to best demonstrate our $U(1)$ theory. One exception is the HCP brain MRI dataset~\cite{VANESSEN201362}, a publicly available set of 1010 subject images acquired with ethics review board approval and patient consent, designed to reflect the demographic diversity of the USA population, including family members and siblings. Family classification from brain MRI is a highly challenging few-shot learning task not yet addressed via deep learning. Handcrafted descriptors have been used to identify images of the same individuals or family members~\cite{chauvin2021efficient}, note subjects are labeled according to random codes with no personal identifiable information. Our experimental pixel vector matching method is effective for classification and demonstrating our theory with general pre-trained CNN architectures, particularly lo-res bottleneck layers, however would require optimisation for larger layers or applications with tight memory or timing constraints. The classification accuracy for models training from scratch is improved by incorporating an antisymmetric $U(1)$ loss function, a result complementary to recent unitary optimization research~\cite{tang2021image,kiani2022projunn}.

\section{Appendices}

\subsection{U(1)-like Symmetry Breaking in the Drosophila Melanogaster Cortex}

We noted that the U(1) symmetry breaking structure in deep CNN bottleneck layers shown in Figure 5 (due to input image class) is similar to examples of angular neural firing patterns observed in deep layers of biological neural networks. Specifically, activations in deep/central brain layers of the drosophila melanogaster cortex linked to the fly’s compass network determining heading~\cite{kim2017ring, drosophilia2015}, in response to dual-photon input stimuli. As central brain regions are typically dominated by recurrent networks which often produce complex patterns of neural activity, it is remarkable to observe a high-level, abstract internal representation in the form of a topological ring representing the heading direction in response to input light stimulus patterns.

\begin{figure}[ht]
\begin{center}
   \includegraphics[width=1\linewidth]{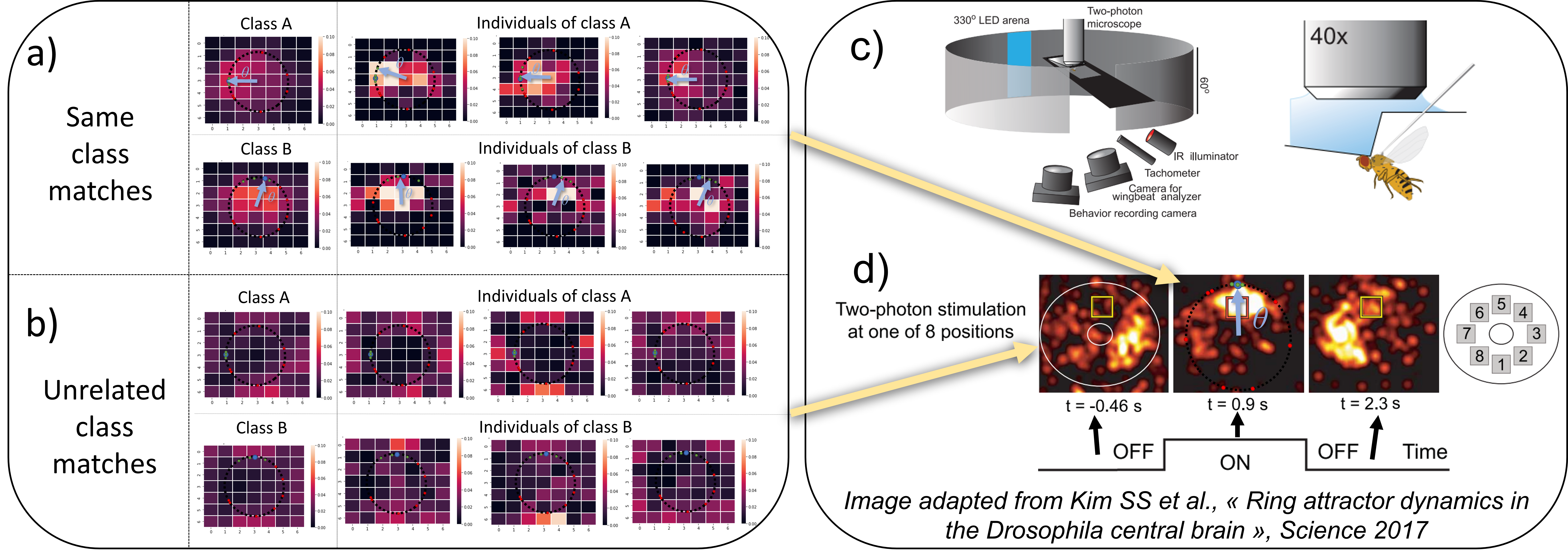} \\
\end{center}
   \caption{Left a) and b) show the $U(1)$ symmetry breaking observed in CNN activations, i.e. NN matches of Densenet201~\cite{huang2016densenet} activation vector layers in response to input Caltech 101~\cite{feifei2006caltech} class images. Matches between same-class images a) exhibit consistent angular deviations $\theta \in U(1)$ for specific classes (e.g. class A, B, blue arrows) and b) unrelated class matches are scattered about the periphery. Right shows the similar responses observed in the central Drosophila cortex following two-photon input stimuli generated in a specialized experimental apparatus ~\cite{kim2017ring} (c). The cortical response (d) exhibits an "activation bump" consistent with a topological ring structure in response to a specific ON stimulus (neural location '5' in the middle picture, similar to our matches to the same class), and a dissipated peripheral structure to OFF stimuli (first and third pictures, similar to our unrelated class matches).}
\label{fig:photons}
\end{figure}

\subsection{Pixel Vector Matching Distributions}

\begin{figure}[ht]
\begin{center}
   \includegraphics[width=1\linewidth]{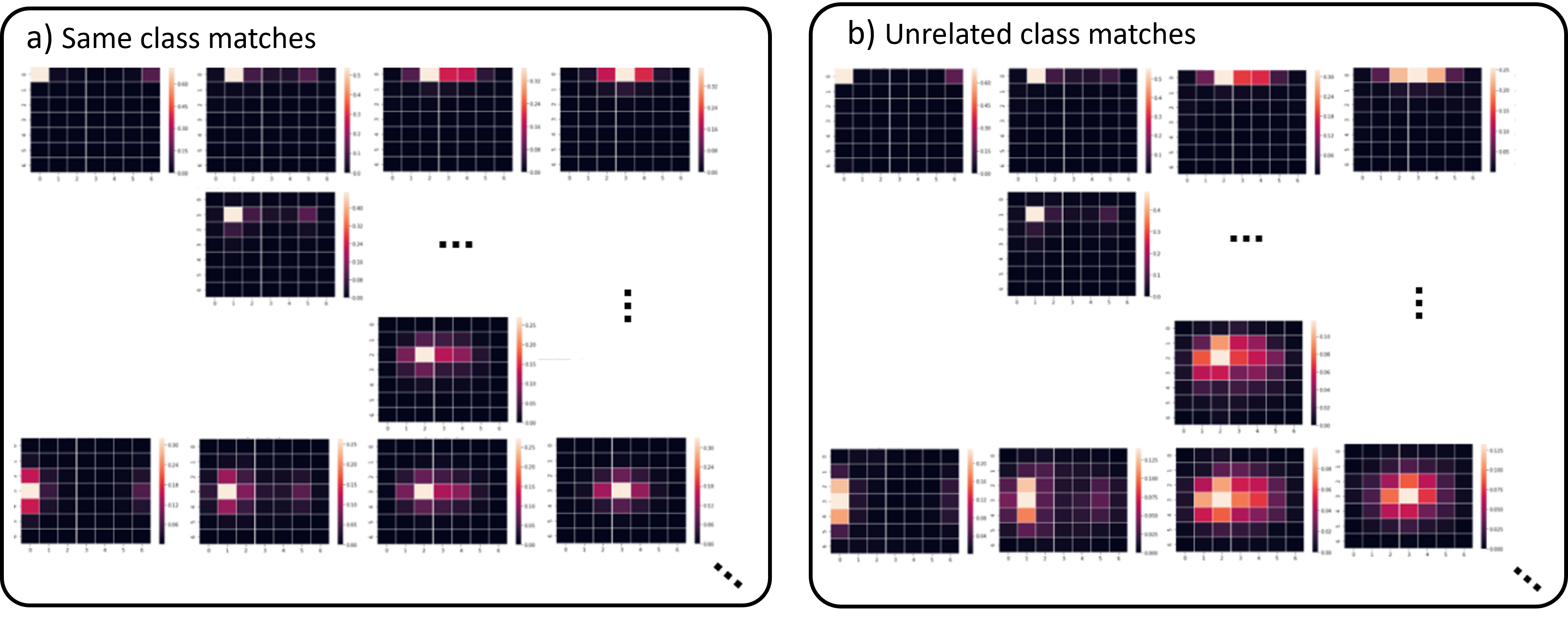} \\
\end{center}
   \caption{Distributions $p(x_{nn}|x_i)$ of matching NN pixel locations $x_{nn}$ conditioned on query locations $x_i$. Memory retrieval is based on 1920-dimensional activation vectors $I_x \in R^{1920+}$ from the $7\times7=49$-pixel bottleneck layer of an Imagenet~\cite{jdeng2009imagenet} pre-trained Densenet201~\cite{huang2016densenet} model and Caltech 101~\cite{feifei2006caltech} class images not used in training. Distributions over $x_{nn}$ are shown for a selection of individual query pixel locations $x_i$. These are tightly clustered around the original query pixel locations $x_i$ (brightest pixels), and exhibit lower variance for matches to instances of the same class (left) vs unrelated class (right). Note that variations are stronger in tangential (as opposed to radial) directions, indicating angular $\theta \in U(1)$ deviations about the unit circle.}
\label{fig:pixelmatching}
\end{figure}

Our experimental observations regarding symmetry are derived from a rudimentary nearest neighbor (NN) classification algorithm (Section 4.1), where activation vectors $I_x$ from individual pixel locations $x$ are matched between query image locations $x_i$ to NN locations $x_{nn}$ of vectors stored in memory. We justify the use of pixel vector indexing by the observation that it generally leads to superior classification performance in comparison to other representations including max pooling or flattening, see Figure 3 and Figure 4, for a number of different networks and classification tasks. Figure 5 shows distributions of nearest neighbor matching locations $x_{nn}$, here Figure~\ref{fig:pixelmatching} shows distributions $p(x_{nn}|x_i)$ conditioned query pixel location $x_i$. Matching locations $x_{nn}$ are tightly distributed around the query pixel locations $x_i$, indicating that in CNN bottleneck layers, activation information $I_{x}$ is highly specific to location $x$, and that unrelated class match distributions exhibit noticeably higher variability in nearest neighbor location $x_{nn}$.

{\small
\bibliographystyle{ieee_fullname}

}

\end{document}